\documentclass[12pt,twoside,a4paper]{article}
\usepackage[OT4]{fontenc}
\usepackage[cp1250]{inputenc}
\usepackage{amsfonts}
\usepackage{amsmath}
\usepackage{amsthm}
\usepackage{graphicx}
\usepackage{hyperref}
\usepackage{float}
\usepackage{url}
\usepackage{array}
\usepackage{algorithm,algpseudocode}

\newcommand{\Bem}[1]{}

\newcommand{\figaddr}[1]{#1}

\newtheorem{theorem}{Theorem}

\begin{document}
\newcommand{\MovjTytulv}{On the Consistency of $k$-means++ algorithm }
\newcommand{\MaInstytucja}{Institute of Computer Science of the Polish Academy of Sciences\\ul. Jana Kazimierza 5, 01-248 Warszawa
Poland}
\title{
\MovjTytulv  }
\author{
Mieczys{\l}aw A. K{\l}opotek (klopotek\@ipipan.waw.pl) 
\\ \MaInstytucja
}
 
\maketitle

\begin{abstract}
We prove in this paper that  the expected value of the objective function of the  $k$-means++ algorithm for samples converges to population expected value. As $k$-means++, for samples,  provides with constant factor approximation for $k$-means objectives, such an approximation can be achieved for the population with increase of the sample size.

This result is of potential practical relevance  when one is considering using subsampling when clustering large data sets (large data bases). 
\end{abstract}

\noindent


\begin{figure}
\centering
\includegraphics[width=0.3\textwidth]{\figaddr{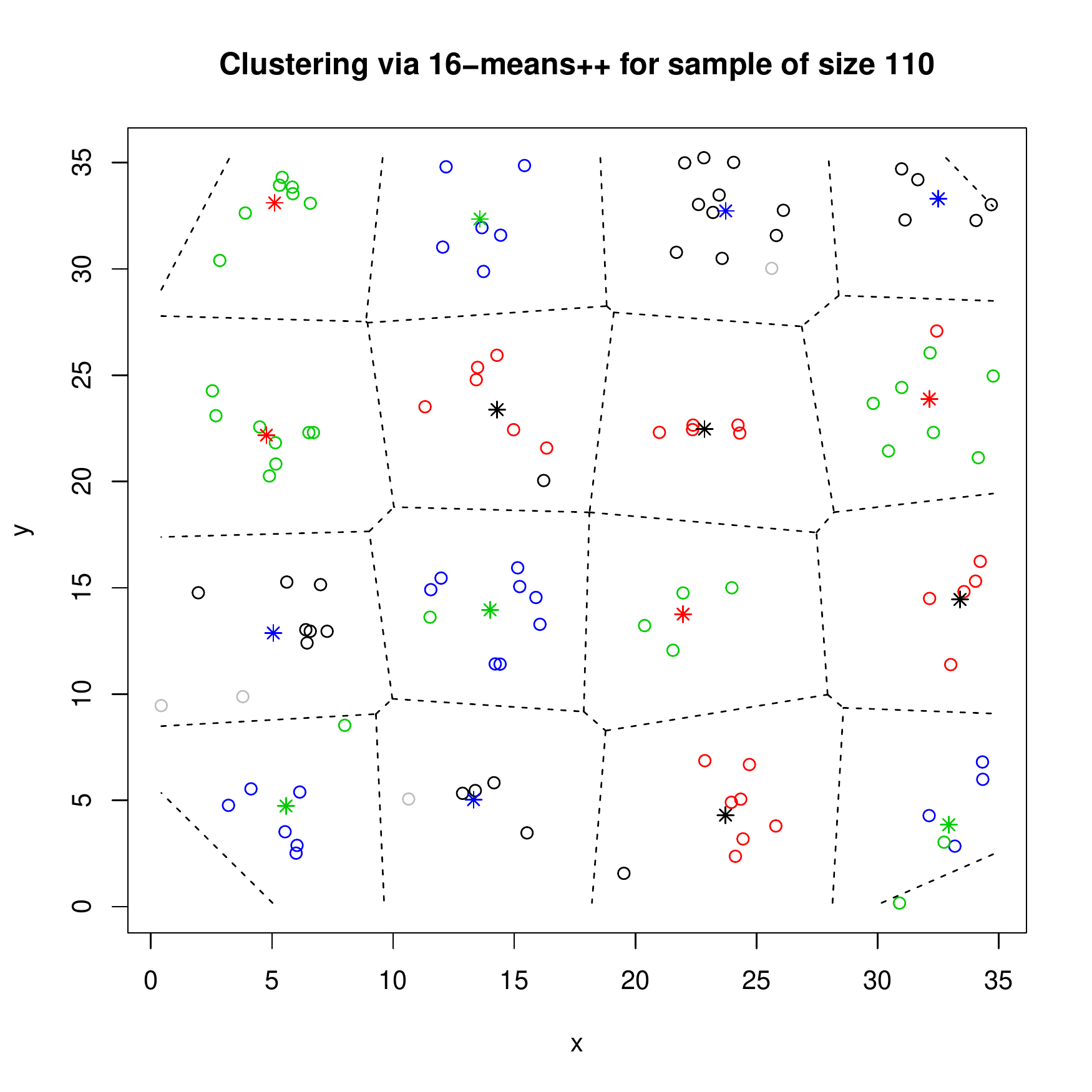}}  %
\includegraphics[width=0.3\textwidth]{\figaddr{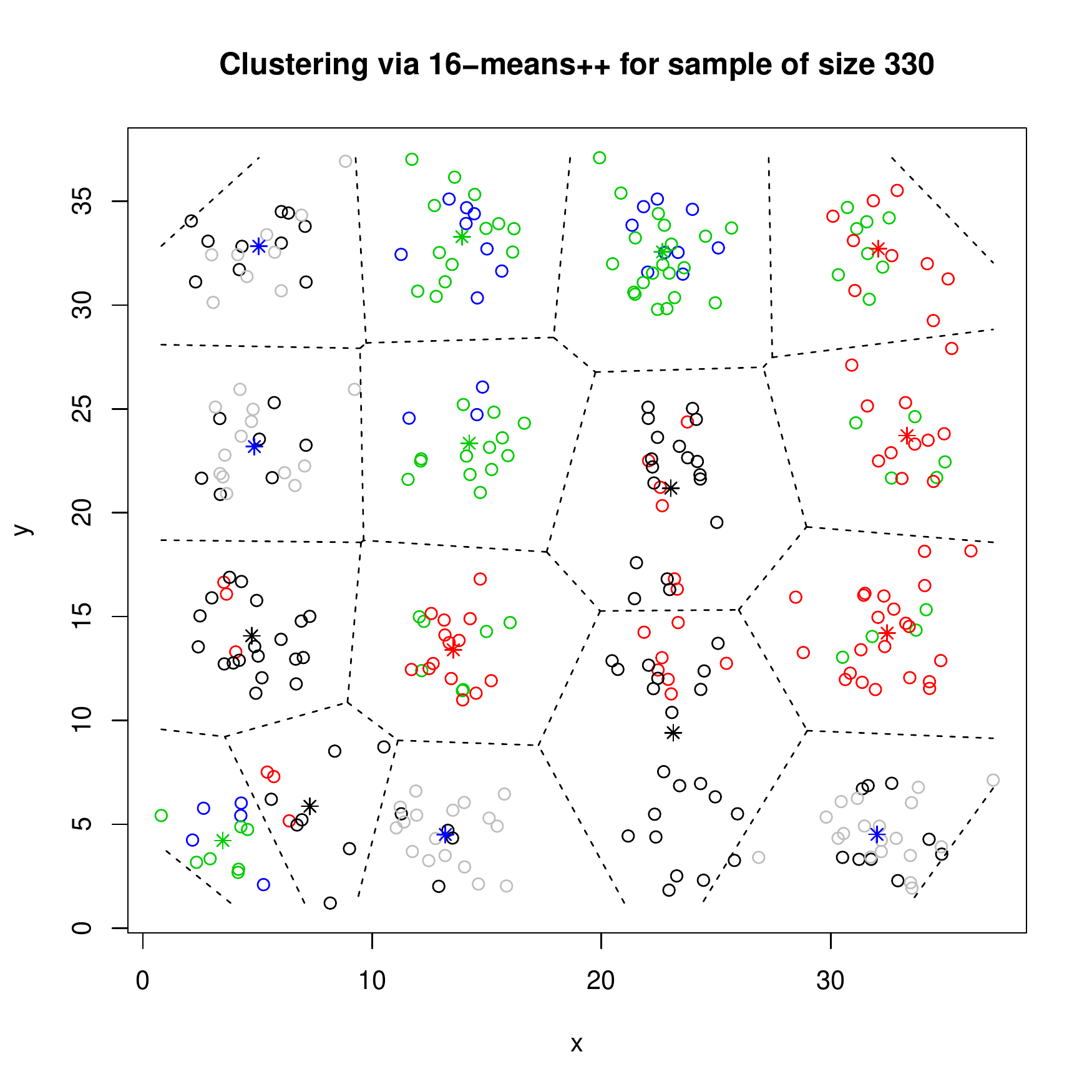}}  %
\includegraphics[width=0.3\textwidth]{\figaddr{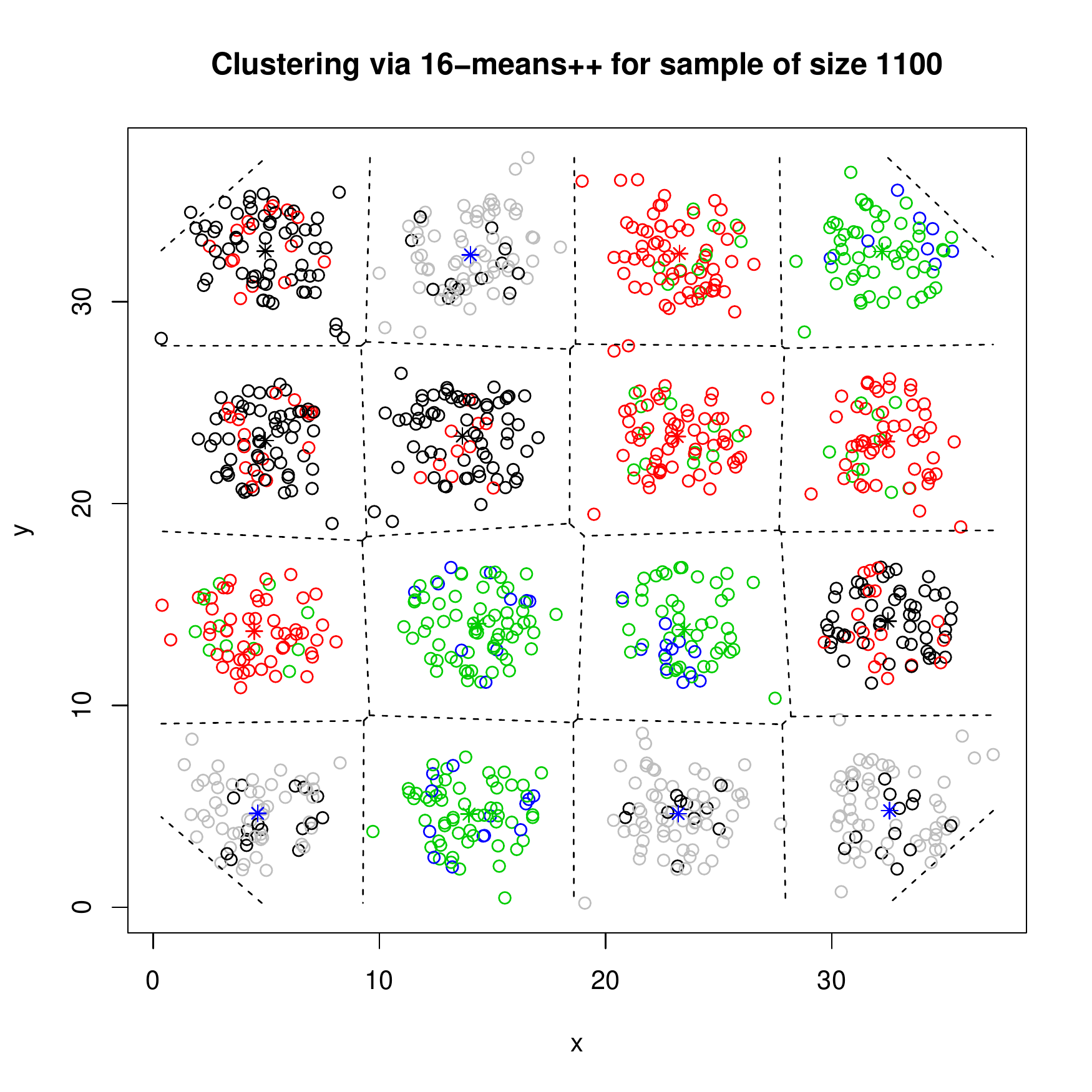}}  %
\\
\includegraphics[width=0.3\textwidth]{\figaddr{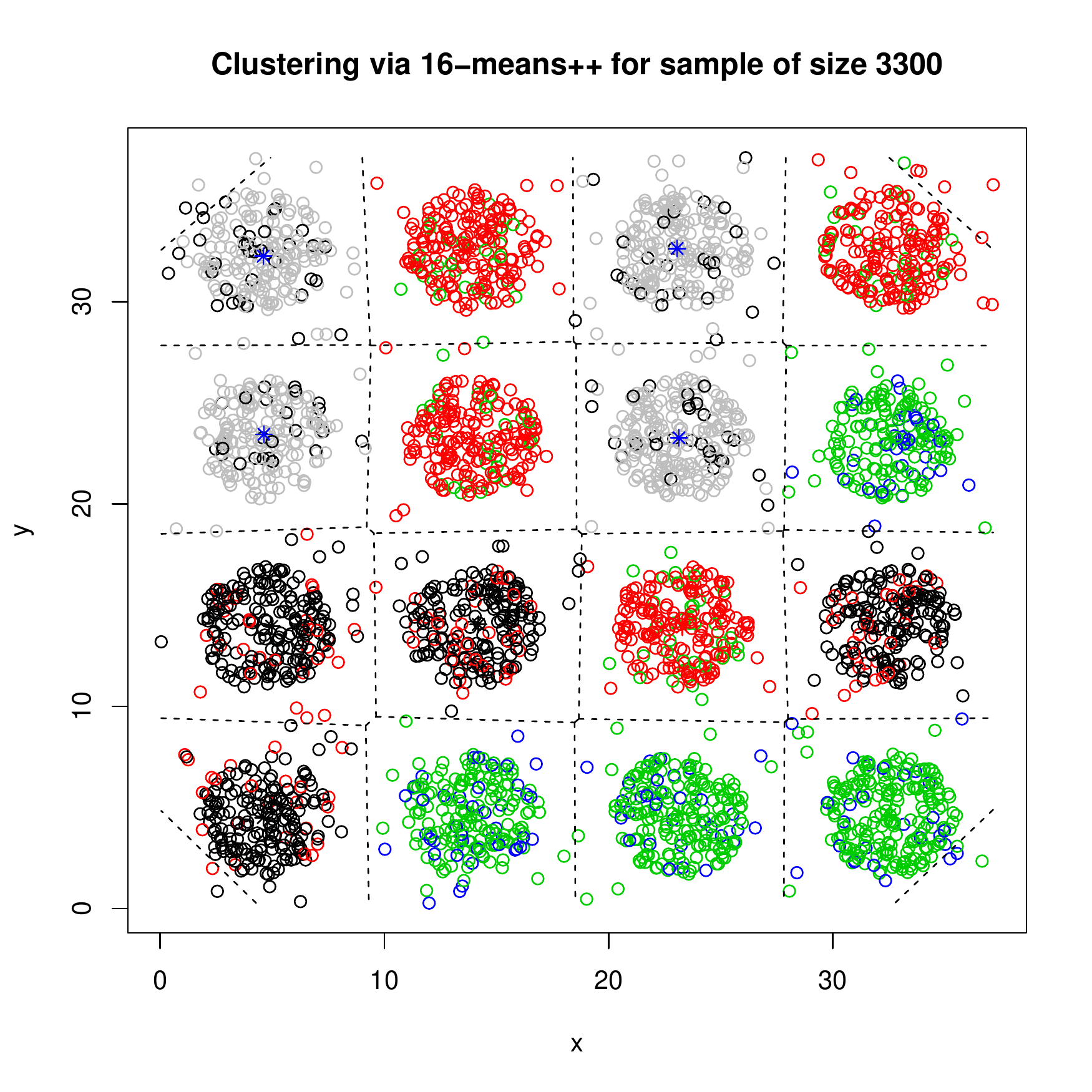}}  %
\includegraphics[width=0.3\textwidth]{\figaddr{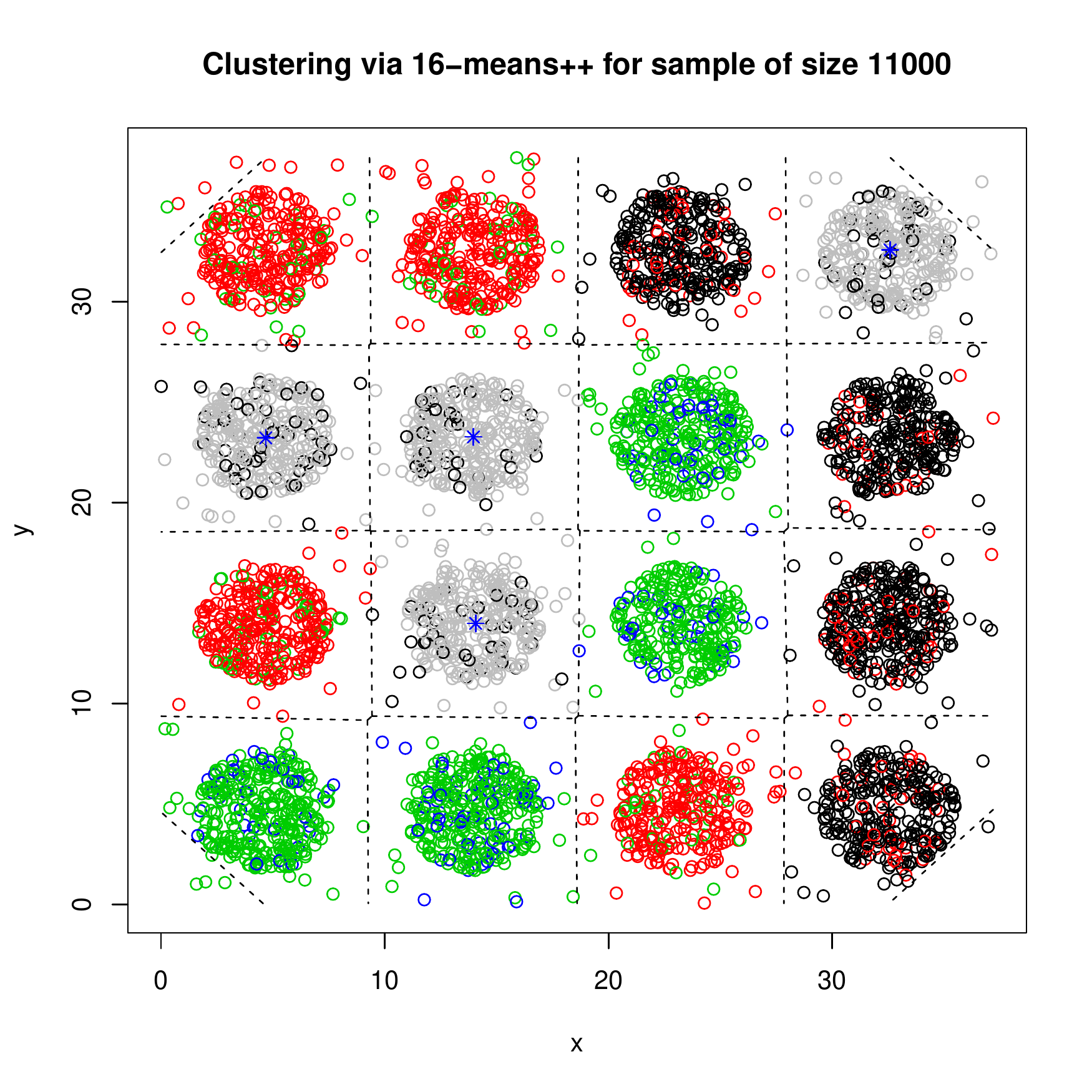}}  %
\includegraphics[width=0.3\textwidth]{\figaddr{rGap16_meanspluplu3300dots.pdf}}  %
\caption{Clustering into 16 clusters of data from the same distribution for various sample sizes as clustered by $k$-means++ algorithm (Voronoi diagram superimposed).  
}\label{fig:kmeansppclusters}
\end{figure}


\begin{figure}
\centering
\includegraphics[width=0.3\textwidth]{\figaddr{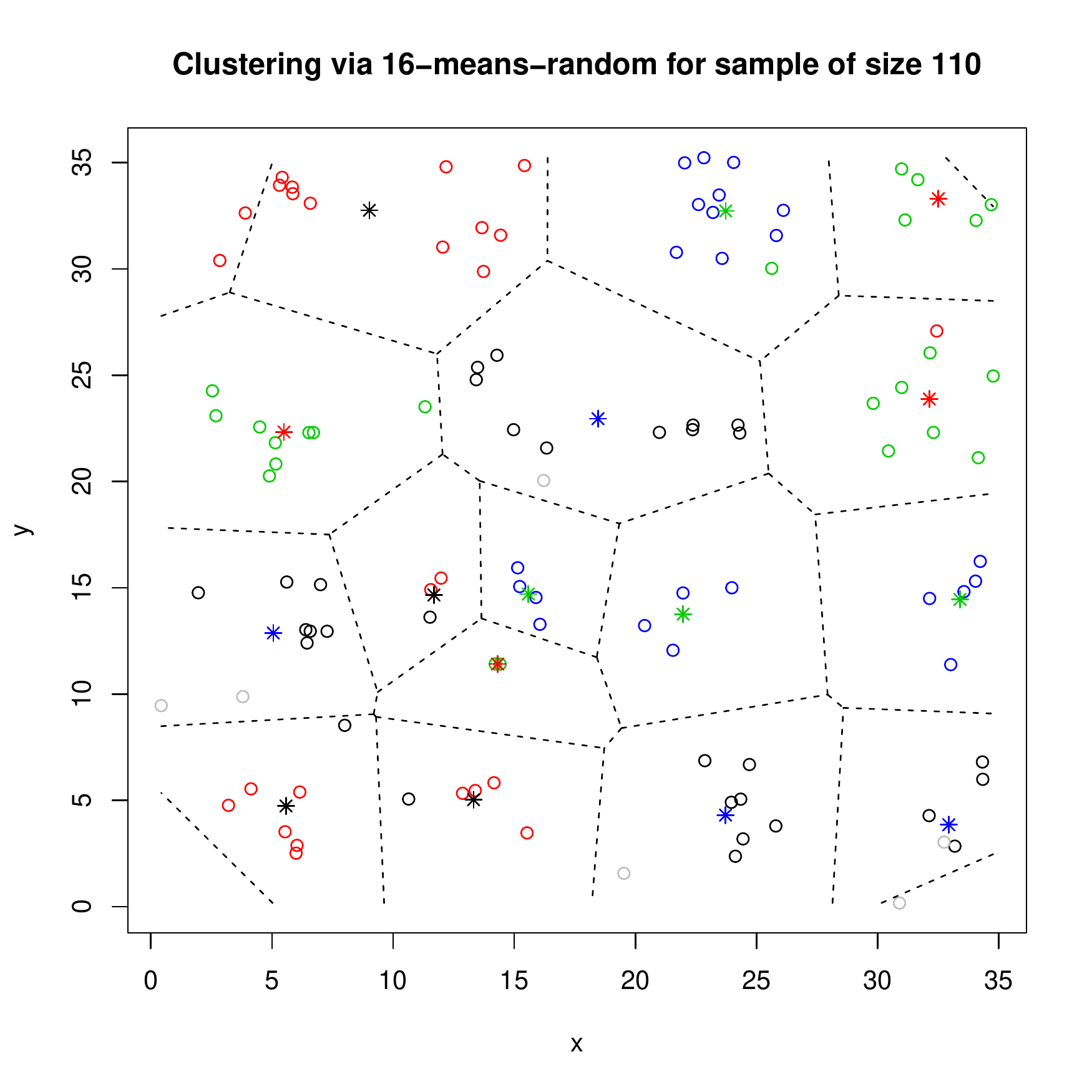}}  %
\includegraphics[width=0.3\textwidth]{\figaddr{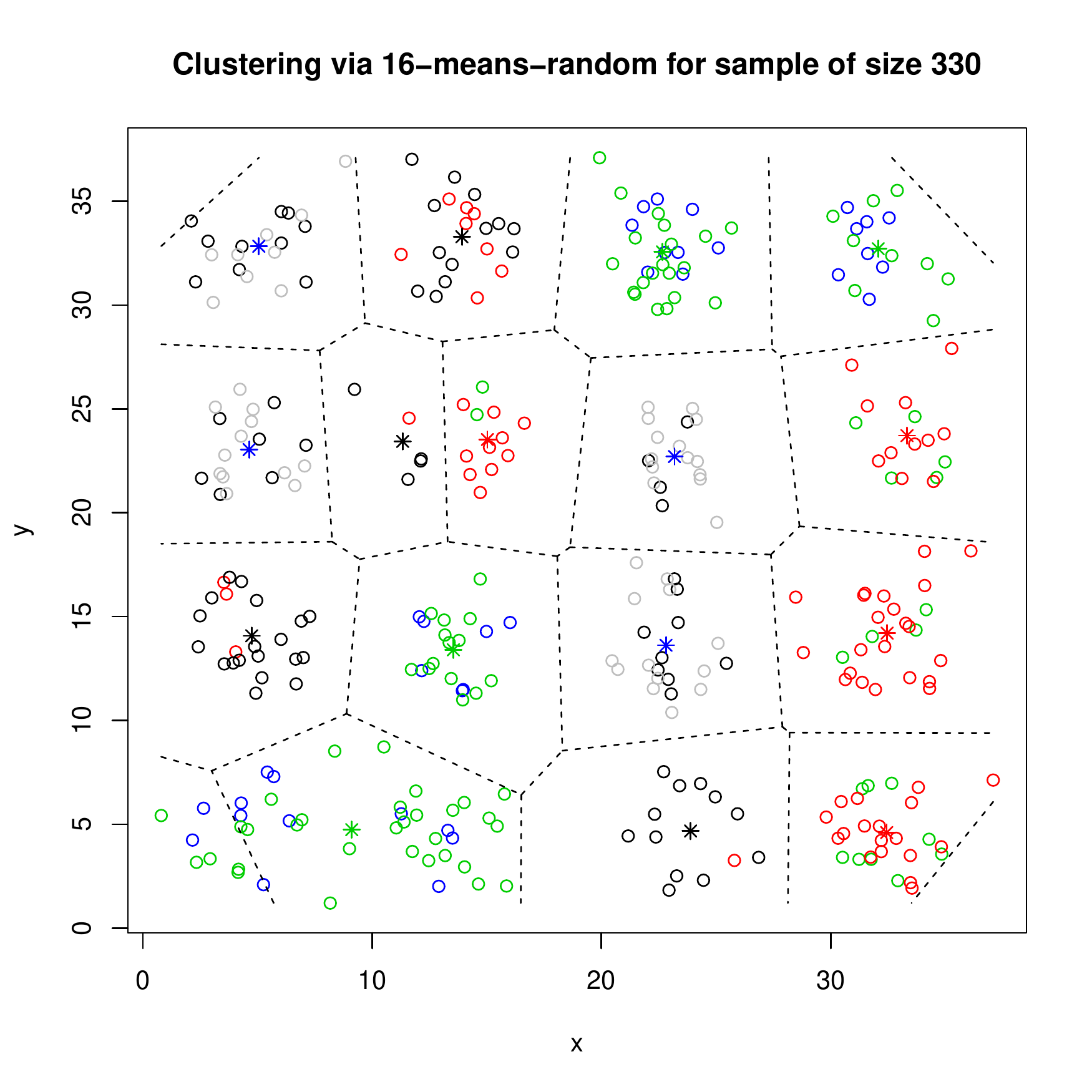}}  %
\includegraphics[width=0.3\textwidth]{\figaddr{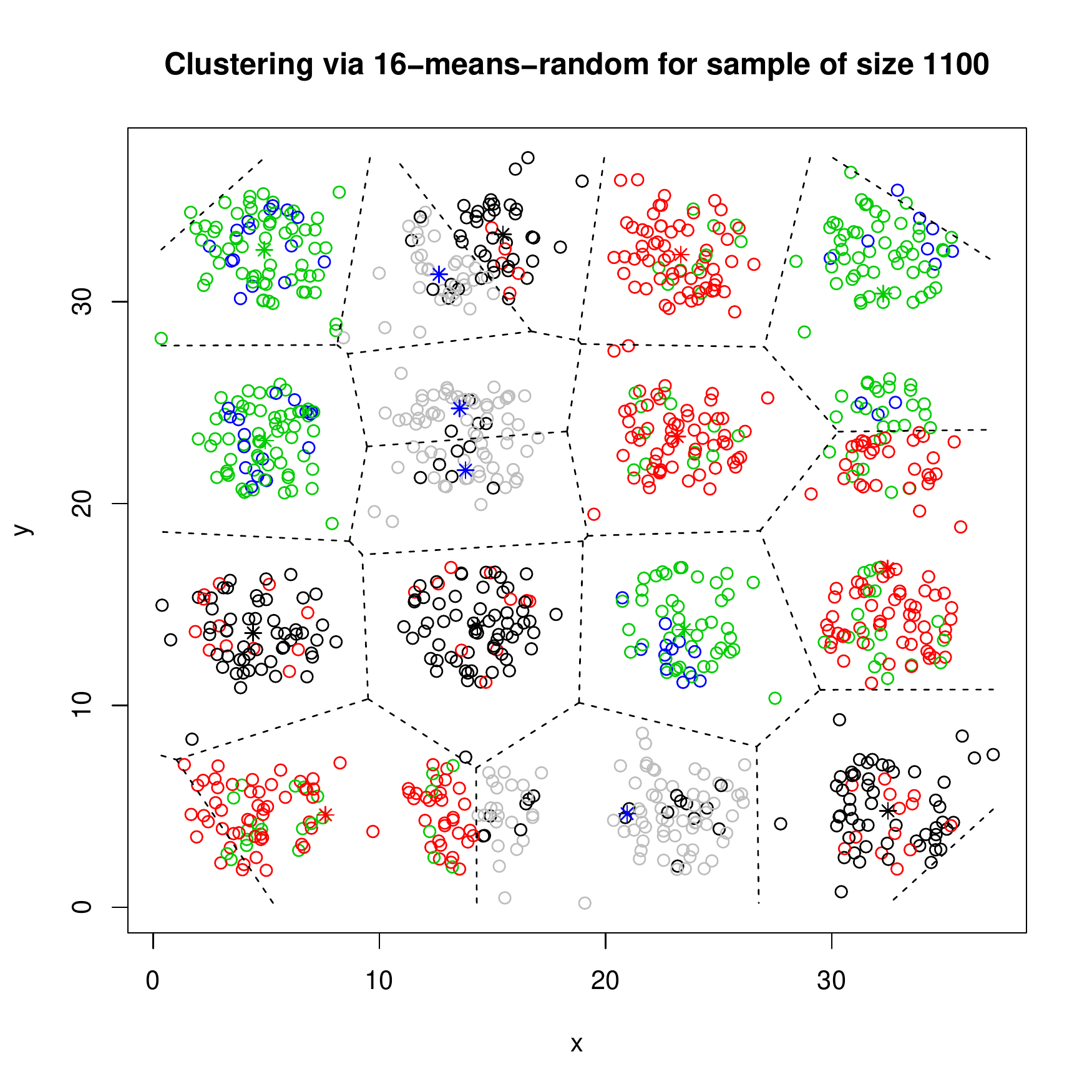}}  %
\\
\includegraphics[width=0.3\textwidth]{\figaddr{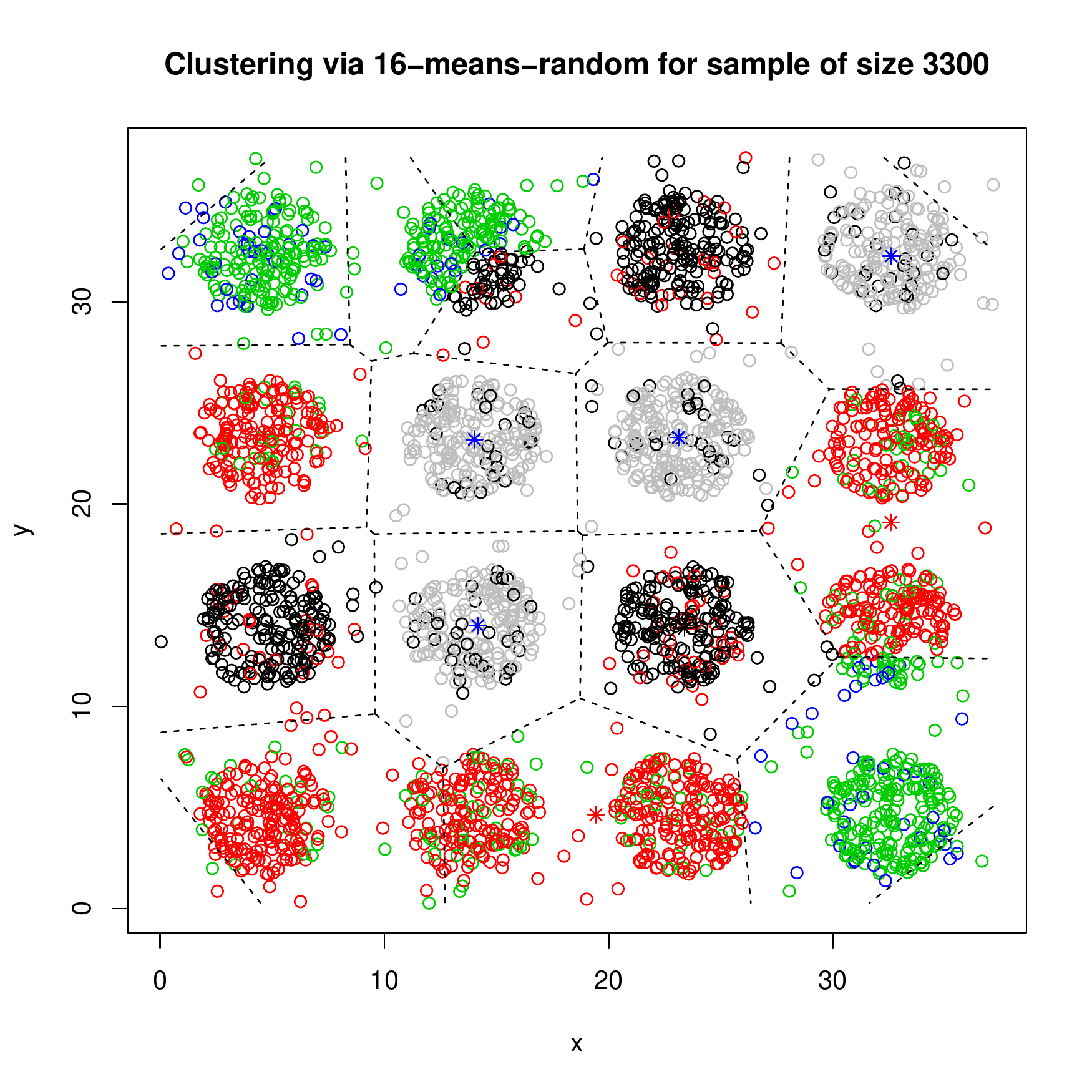}}  %
\includegraphics[width=0.3\textwidth]{\figaddr{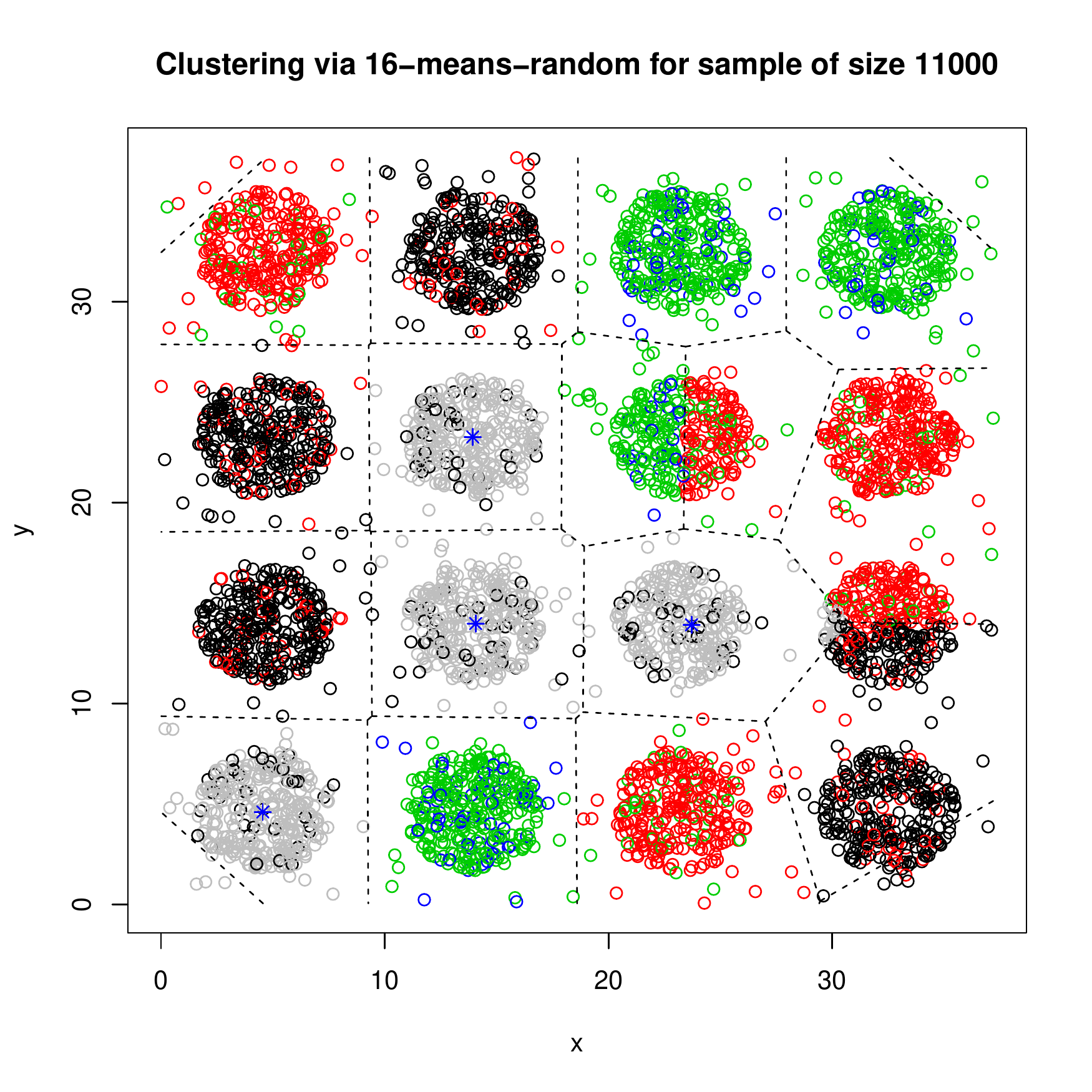}}  %
\includegraphics[width=0.3\textwidth]{\figaddr{rGap16_meansrandom3300dots.pdf}}  %
\caption{Clustering into 16 clusters of data from the same distribution for various sample sizes as clustered by $k$-means-random algorithm (Voronoi diagram superimposed).  
}\label{fig:kmeansrandomclusters}
\end{figure}

\section{Introduction}

The common sense feeling about data mining algorithms is that with increasing sample sizes the tasks should become conceptually easier though more though computationally. 

One of these areas is surely the cluster analysis. 
A popular clustering algorithm, $k$-means, shall be studied in this research with respect to its behaviour "in the limit". 
It essentially strives to minimise the partition quality function (called also "partition cost function") 

\begin{equation}\label{CLU:obj-1}
J(U,M)=\sum_{i=1}^m\sum_{j=1}^k u_{ij}\|\textbf{x}_i - \boldsymbol{\mu}_j\|^2
\end{equation}
\noindent 
where 
$\textbf{x}_i$, $i=1,\dots, m$ are the data points, 
$M$ is the matrix of cluster centres 
$ \boldsymbol{\mu}_j$, $j=1,\dots, k$, and 
$U$ is the cluster membership indicator matrix, consisting of entries $u_{ij}$,
where 
$u_{ij}$ is equal 1 if among all of cluster centres $\boldsymbol{\mu}_j$ is the closest to 
$\textbf{x}_i$, and is 0 otherwise. 

$k$-means
  comes in various variants, most of them differing by the way how the initial hypothetical cluster centres are found.
We will concentrate on the so-called $k$-means++, where the initialisation is driven by candidate probability of being selected proportional to squared distances to the closest cluster centre candidate selected earlier.
This distinguishes it from the basic version, called here $k$-means-random, where the initial candidates are selected randomly with uniform probability distribution over data set. 
Both are described in detail in the paper \cite{CLU:AV07} by Arthur and Vassilvitskii.
 
In Figure \ref{fig:kmeansppclusters} you see the results of clustering for data from a relatively easy probability distribution of 16 identical clusters arranged uniformly in 4 rows of 4 clusters. 
In Figure \ref{fig:kmeansrandomclusters} you see the results of clustering for exactly the same data samples. 
It is immediately visible, that even in simple cases the algorithms, for small samples, do not return clusters identical with those for the "in the limit" distribution.
In particular, $k$-means-random behaves poorly even for quite large samples (of 33000 data points).
$k$-means++ seems to stabilize with sample size, but over multiple repetitions bad cluster split can occur.

The behaviour "in the limit" is of practical relevance, as   
some researchers, especially in the realm of database mining, 
propose to cluster "sufficiently large" samples instead of the whole database content.
See for example  Bejarano et al. \cite{Bejarano:2011} on $k$-means accelerating via subsampling. 

For these reasons Pollard \cite{Pollard:1981} investigated already in 1981 the “in the limit” behaviour of $k$-means, while MacQueen \cite{MacQueen:1967}  considered it already in 1967. 

The essential problem consists in the question whether or not we can expect that the cost function value from equation (\ref{CLU:obj-1}),  and hence the "best clustering", of the population  can be approximated by the results of clustering samples (of increasing sizes).

Regrettably, researchers considered in the past an unrealistic version of $k$-means, that we shall call here "$k$-means-ideal", that is able to find the minimum of (\ref{CLU:obj-1}) for a sample which is $NP$-hard.
This tells nothing about the real-world algorithms, like $k$-means-random or $k$-means++, because they are not guaranteed to approach this optimum in any way. 
In fact, no results for $k$-means-random are known even for constant factor approximations to sample optimal value. 
So far, only for $k$-means++ we have this kind of approximation provided by Arthur and Vassilvitskii \cite{CLU:AV07} . 

The constant factor approximation for a sample does not guarantee the existence of the constant factor approximation of optimal cost function value for the population as the limit of sample constant factor approximations,
because the whole proof of in the limit behaviour of $k$-means-ideal  by Pollard   relies on the assumption of the existence of unique optimal solution. 
No direct transposition is possible to any constant factor approximation because there exist potentially infinite many constant factor approximation  hence also as many constant factor approximating partitions.

So in this paper we address the issue whether or not it is possible to 
get a constant factor approximation clustering of the population 
as a limit of constant factor approximation partitions of the samples with increasing sample sizes.

\section{Previous work}

Arthur and Vassilvitskii \cite{CLU:AV07}   proved the following property of their $k$-means++ algorithm.

\begin{theorem}\cite{CLU:AV07} The expected value of the partition cost, computed according to equation  (\ref{CLU:obj-1}), is delimited in case of  $k$-means++ algorithm by the inequality
\begin{equation}
\mathbb{E}(J) \le 8(\ln k + 2)J_{opt}
\end{equation}
\noindent where $J_{opt}$ denotes the optimal value of the partition cost.   
\end{theorem}

(For $k=3$ this amounts to more than 24). 
Note that the proof in  \cite{CLU:AV07} refers to the initialisation step of the algorithm only so that subsequently we assume that $k$-means++ clustering consists solely of this initialisation. 
In the full algorithm $\mathbb{E}(J) $ is lower than after initialisation step only.

There exist a number of other constant factor approximation algorithms for $k$-means objective  beside  \cite{CLU:AV07}.
But they either decrease the constant below $8(\ln k + 2)$ at the expense of an excessive complexity, still with constant factor above 9 \cite{Kanungo:2002},   
by restricting to well-separated cases \cite{Tang:2015}, 
or by approximating the cost function of $k$-means with $k+l$-means, where $l$ is a quite large number, see e.g. the paper   \cite{Sen:2004}.   
Still most of them rely on the basic approach of  \cite{CLU:AV07}.
Other approaches, like that of \cite{Balcan:2009}, assume that \emph{all} the near optimal partitions lie close to the optimal solution without actually proving it.

All this research attempts to annihilate the fundamental problem behind the $k$-means-random, namely, as    \cite{CLU:AV07} states:
"It is the speed and simplicity of the $k$-means[-random] method that make it appealing, not its accuracy.
Indeed, there are many natural examples for which the algorithm generates arbitrarily bad clusterings
(i.e.,$\frac{J}{J_{opt}}$ is unbounded even when $m$ and $k$ are fixed)."

While the aforementioned research direction attempts to improve sample-based cost function results for $k$-means, 
 there existed and exists interest in strong consistency results concerning variants of $k$-means algorithm, aiming at proving properties achieved with increasing sample size. 
Pollard \cite{Pollard:1981}
proved an interesting property of a hypothetical $k$-means-ideal  algorithm 
that would produce a partition yielding $J_{opt}$ for finite sample.  
He demonstrated, under assumption of a unique global optimum, the convergence of $k$-means-ideal  algorithm to proper  population clustering with increasing sample size. 
  So one can state that such an algorithm would be strongly consistent. 
MacQueen \cite{MacQueen:1967} considered a $k$-means-ideal algorithm under the settings where there exists no unique clustering reaching the minimum of the cost function for the population.
Hence $k$-means-ideal may be unstable for the sequence of samples of increasing size. 
Terada \cite{Terada:2014} considered a version of $k$-means called "reduced" $k$-means which seeks a lower dimensional subspace in which $k$-means can be applied.
He again assumes that a $k$-means-ideal version is available when carrying out his investigations.
The problem of the $k$-means-ideal algorithm is that it is $NP$-hard \cite{Pollard:1981},  even in 2D \cite{Mahajan:2009}).

In summary, in the mentioned works   an abstract global optimiser is referred to rather than an actual practically used algorithm.  

Therefore in this research we study the in the limit behaviour of the realistic $k$-means++ algorithm. 
Though it is slow compared to $k$-means-random, various accelerations have been proposed, e.g. \cite{Bachem:2016}, hence it is with studying.  

\section{In the limit behaviour of $k$-means++}

So  let us investigate whether or not a realistic algorithm, 
$k$-means++, featured by reasonable complexity and reasonable expectation of closeness to optimality, is also in some sense strongly consistent.

In what follows we extend the results of Pollard. 

\subsection{Notation and assumption}
Assume that we are drawing independent samples of size $m=k,k+1,\dots .$ from some probability distribution $P$. 
Assume further, that the number of disjoint balls of radius $\epsilon_b$ for some  $\epsilon_b>0$ with non-zero $P$ measure   is much larger than $k$. 
Furthermore let $$\int \|\textbf{x}\|^{2(k+1)} P(d\textbf{x})$$ be finite. 

Given a sample consisting of  $\textbf{x}_i$, $i=1,\dots ,m$, it will be convenient to consider  an empirical probability distribution $P_m$ assigning a probability mass of $m^{-1}$ at each of the sample points $\textbf{x}_i$.
Let further $M$ denote any set of up to $k$ points (serving as an arbitrary set of cluster centres).

Let us introduce the function $J_{cp}(.,.)$, that  for any set $M$ of points in  
$\mathbb{R}^n$ 
and any probability distribution $Q$ over $\mathbb{R}^n$,
computes $k$-means quality function normalised over the probability distribution: 
\begin{equation} 
J_{cp}(M, Q)=\int  D(\textbf{x} ,M)^2 Q(d\textbf{x})
\end{equation}
where $D(\textbf{x} ,M)=\min_{\boldsymbol\mu \in M}  \|\textbf{x}  - \boldsymbol{\mu}\|$.

Then $J_{cp}(M, P_m)$ can be seen as a version  of the function $J()$ from equation (\ref{CLU:obj-1}).

Note that for any sets $M. M'$ such that $M\subset M'$ 
we have $J_{cp}(M, Q)\ge J_{cp}(M', Q)$. 
 
Note also that if $M$ has been obtained via the initialisation step of $k$-means++, then 
the elements of $M$ can be considered as ordered, 
so we could use indexes $M_{[p:q]}$ meaning elements $\boldsymbol\mu_p,\dots ,\boldsymbol\mu_q$ from this sequence. 
As the cluster around $M_{[k]}$ will contain at least one element ($\boldsymbol\mu_k$ itself), 
we can be sure that 
 $J_{cp}(M_{[1:k-1]}, P_m)> J_{cp}(M_{[1:k]}, P_m)$ as no element has a chance to be selected twice. 
Hence also 
$\mathbb{E}_M(J_{cp}(M_{[1:k-1]}, P_m)) > \mathbb{E}_M(J_{cp}(M_{[1:k]}, P_m))$.


\Bem{
Note that Pollard has proven that 
for a fixed set of centres $M$, by the virtue of the strong low of  large numbers
\begin{equation} 
J_{cp}(M, P_m) \rightarrow J_{cp}(M, P ) =\int  \min_{\boldsymbol\mu \in M}  \|\textbf{x}  - \boldsymbol{\mu}\|^2 P(d\textbf{x})
\end{equation}
}

\subsection{Main result}

If we assume that the distribution $P$ is continuous, 
then for the expectation over all possible  $M$ obtained from initialisation procedure of $k$-means++, 
 for any $j=2,\cdots k$ the relationship 
$\mathbb{E}_M(J_{cp}(M_{[1:j-1]}, P)) > \mathbb{E}_M(J_{cp}(M_{[1:j]}, P))$ holds.  
The reason is as follows:
The last point, $\boldsymbol\mu_k $, was selected from a point of non-zero density, so that in ball 
of radius $\epsilon_b>0$ ($\epsilon_b$ lower than 1/3 of the distance of $\boldsymbol\mu_j$ to each  $\boldsymbol\mu \in M_{[1:j-1]}$) around it 
would be re-clustered to $\boldsymbol\mu_j$  diminishing the overall within-cluster variance. 
As it holds for each set $M$, so it holds for the expected value too. 

\emph{
But our goal here is to show that if we pick the sets $M$ according to 
$k$-means++ (initialization) algorithm, then 
$\mathbb{E}_M(J_{cp}(M, P_m))\rightarrow_{m\rightarrow \infty} 
 \mathbb{E}_M(J_{cp}(M, P))$, and, as a consequence, 
$\mathbb{E}_M(J_{cp}(M, P)) \le 8(\ln k + 2)J_{.,opt}$.
}

We show this below.
\Bem{
We demonstrate that the probability distribution of $M$ selected 
by the $k$-means++ algorithm for the distribution $P_m$ 
converges to that for $P$. 
If it is so, so the expected value does. 
}

Let $T_{m,k}$ denote the probability distribution for choosing the set $M$ 
as cluster centres for a sample from $P_m$ using $k$-means++ algorithm. 
From the $k$-means++ algorithm we know that 
  the probability of choosing $\boldsymbol\mu_j$ 
given $\boldsymbol\mu\in M_{[1:j-1]}$ have already been picked amounts to 
$Prob(\boldsymbol\mu_j|M_{[1:j-1]})=
\frac{ D( \boldsymbol\mu_j, M_{[1:j-1]})^2 }{\int  D(\textbf{x}, M_{[1:j-1]})^2 P_m(d\textbf{x})}$.
whereas the point $\boldsymbol\mu_1$ is picked according to the probability distribution $P_m$. 
Obviously , the probability of selecting the set $M$ amounts to: 
$$Prob(M)= Prob(\boldsymbol\mu_1)
\cdot Prob(\boldsymbol\mu_2|M_{[1]})
\cdot Prob(\boldsymbol\mu_3|M_{[1:2]})
\cdot \dots \cdot Prob(\boldsymbol\mu_k|M_{[1:k-1]})$$.

So the probability density $T_{m,k}$ may be expressed as 
$$P_m(\boldsymbol\mu_1) \cdot \prod_{j=2}^k \frac{D(  \boldsymbol\mu_j,  M_{[1:j-1]})^2 }{\int  D(\textbf{x}, M_{[1:j-1]})^2  P_m(d\textbf{
x})}
$$
For $P$ we get accordingly the $T_{k}$ distribution.

Now assume we ignore with probability of at most $\delta>0$ a “distant” part of the potential locations of cluster centres from $T_k$ via a procedure described below. 
Denote with $\textbf{M}_\delta$ the subdomain of the set of cluster centres left after ignoring these elements. 
Instead of expectation of 
$$\mathbb{E}_M(J_{cp}(M,P_m) )
= \int J_{cp}(M,P_m)T_{m,k}(dM)
$$
we will be interested in 
$$\mathbb{E}_{M,\delta}(J_{cp}(M,P_m) )
= \int_{M\in \textbf{M}_\delta}  J_{cp}(M,P_m)T_{m,k}(dM)
$$
and respectively for $P$ instead of 
$$\mathbb{E}_M(J_{cp}(M,P) )
= \int J_{cp}(M,P)T_{k}(dM)
$$
we want to consider 
$$\mathbb{E}_{M,\delta}(J_{cp}(M,P ) )
= \int_{M\in \textbf{M}_\delta}  J_{cp}(M,P )T_{ k}(dM)
$$

$ \mathbb{E}_M(J_{cp}(M, P))$ is apparently finite. 

Let us concentrate on the difference
$|\mathbb{E}_{M,\delta}(J_{cp}(M,P ) )
- \mathbb{E}_{M,\delta}(J_{cp}(M,P_m ) )|
$ 
and show that it converges to 0 when $m$ is increased for a fixed $\delta$, 
  that is for any $\epsilon$ and $\delta$ 
 there exists such an $m_{\delta, \epsilon} $
 that for any larger $m$ almost surely (a.s.) this difference is lower than $\epsilon$. 
Thus by decreasing $\delta$ and $\epsilon$ we find that the expectation on $P_m$ converges to that on $P$. 

The subdomain selection (ignoring $\delta$ share of probability mass of $T_m$) shall proceed as follows.
Let $R_1$ be a radius such that 
$\int_{\|\textbf{x}\|>R_1} \|\textbf{x}\|^2 P(d\textbf{x}) <\delta/k$. 
We reject all $M$ such that $\|\boldsymbol\mu_1  \|>R_1$. 

Let us denote 
$J_{m,opt}= \min_M J_{cp}(M, P_m)$, 
$J_{.,opt}= \min_M J_{cp}(M, P)$, 
where $M$ is a set of cardinality at most $k$.
Let us denote 
$J_{m,opt,j}= \min_M J_{cp}(M, P_m)$, 
$J_{.,opt,j}= \min_M J_{cp}(M, P)$,
where $M$ be a set of cardinality at most $j$, $j=1,\dots ,k$.

\newcommand{\bmu}{\boldsymbol\mu}

Now observe for $l=2,\dots,k$ that 
$$
\int _{\bmu_1} \dots \int _{\bmu_l}  
\prod_{j=2}^{l} \frac{D(  \bmu_j,  M_{[1:j-1]})^2 }{\int  D(\textbf{x}, M_[{1:j-1]})^2  P(d\textbf{
x})}
  P(d \bmu_l)\dots P(d\bmu_1 )
$$
$$
\le 
\int _{\bmu_1} \dots \int _{\bmu_l}  
\prod_{j=2}^l \frac{D(  \bmu_j,  M_{[1:j-1]})^2 }{J_{.,opt,j-1}}
  P(d \bmu_l)\dots P(d\bmu_1 )
$$
$$
= \left(\prod_{j=1}^{l-1} \frac1{J_{.,opt,j}}\right)
\int _{\bmu_1} \dots \int _{\bmu_l}  
\prod_{j=2}^l  D(  \mu_j,  M_{[1:j-1}])^2 
  P(d \bmu_l)\dots P(d\bmu_1 )
$$
Upon restricting the integration area on $\bmu_1,\dots,\bmu_{l-1}$ 
 $$
\le \left(\prod_{j=1}^{l-1} \frac1{J_{.,opt,j}}\right)
\int _{\|\bmu_ 1 \|\le R_1} \dots  \int _{\|\bmu_{l-1}\|\le R_{l-1}}\int _{\bmu_l }
\prod_{j=2}^l  D(  \mu_j,  M_{[1:j-1}])^2 
  P(d \bmu_l)\dots P(d\bmu_1 )
$$
Now substituting the maximum distance between points $\bmu_1,\dots,\bmu_{l-1}$
 $$
\le \left(\prod_{j=1}^{l-1} \frac1{J_{.,opt,j}}\right)
\int _{\|\bmu_ 1 \|\le R_1} \dots  \int _{\|\bmu_{l-1}\|\le R_{l-1}}\int _{\bmu_l }\left(\prod_{j=2}^{l-1}   (R_j+R_j-1)^2 \right)
 D(  \mu_l,  M_{[1:l-1}])^2 
$$ $$  P(d \bmu_l)\dots P(d\bmu_1 )
$$
 $$
= \left(\prod_{j=1}^{l-1} \frac1{J_{.,opt,j}}\right)
\left(\prod_{j=2}^{l-1}   (R_j+R_j-1)^2 \right)
\int _{\|\bmu_ 1 \|\le R_1} \dots \int _{\|\bmu_{l-1}\|\le R_{l-1}}\int _{\bmu_l } D(  \mu_l,  M_{[1:l-1}])^2 
$$ $$  P(d \bmu_l)\dots P(d\bmu_1 )
$$
Now substituting the maximum distance to the last point $\bmu_l$ 
 $$
\le \left(\prod_{j=1}^{l-1} \frac1{J_{.,opt,j}}\right)
\left(\prod_{j=2}^{l-1}   (R_j+R_j-1)^2 \right)
\int _{\|\bmu_ 1 \|\le R_1} \dots \int _{\|\bmu_{l-1}\|\le R_{l-1}}\int _ (\|  \mu_l \|+R_{l-1} )^2 
$$ $$  P(d \bmu_l)\dots P(d\bmu_1 )
$$
 $$
\le \left(\prod_{j=1}^{l-1} \frac1{J_{.,opt,j}}\right)
\left(\prod_{j=2}^{l-1}   (R_j+R_j-1)^2 \right)
 \int _{\bmu_l }
 (\|  \mu_l \|+R_{l-1} )^2 
  P(d \mu_l)
$$
Let us choose now  an $R_l$ so that 
$$\left(\prod_{j=1}^{l-1} \frac1{J_{.,opt,j}}\right)
\left(\prod_{j=2}^{l-1}   (R_j+R_j-1)^2 \right)
 \int _{\|\mu_l\| \le R_l}
 (\|  \mu_l \|+R_{l-1} )^2 
  P(d \mu_l) <\delta/k$$

In this way we again reject at most $\delta/k$ $\mu_l$ points.
Upon repeating this process for $l=2,\dots,k$,   we will reject at most $\delta$ share of potential  sets of centres under  $T_k$.

So let us consider 

$$
|\mathbb{E}_{M,\delta}(J_{cp}(M,P ) )
- \mathbb{E}_{M,\delta}(J_{cp}(M,P_m ) ) |
$$ 
$$
=
| \int_{M\in \textbf{M}_\delta}  J_{cp}(M,P_m)T_{m,k}(dM)
 - \int_{M\in \textbf{M}_\delta}  J_{cp}(M,P )T_{ k}(dM) 
| 
$$
$$
=
| \int_{M\in \textbf{M}_\delta}  J_{cp}(M,P_m)
\prod_{j=2}^{k} \frac{D(  \mu_j,  M_{[1:j-1}])^2 }{\int  D(\textbf{x}, M_{[1:j-1}]^2  P_m(d\textbf{
x})}
  P_m(d \bmu_k) \dots P_m(d \bmu_1)
$$ $$%
 - \int_{M\in \textbf{M}_\delta}  J_{cp}(M,P ) 
\prod_{j=2}^{k} \frac{D(  \mu_j,  M_{[1:j-1}])^2 }{\int  D(\textbf{x}, M_{[1:j-1}]^2  P(d\textbf{
x})}
  P(d \bmu_k)\dots P(d\bmu_1 )
| 
$$
$$
\le
| \int_{M\in \textbf{M}_\delta}  J_{cp}(M,P_m)
\prod_{j=2}^{k} \frac{D(  \mu_j,  M_{[1:j-1}])^2 }{J_{m,opt,j-1}}
  P_m(d \bmu_k) \dots P_m(d \bmu_1)
$$ $$%
 - \int_{M\in \textbf{M}_\delta}  J_{cp}(M,P ) 
\prod_{j=2}^{k} \frac{D(  \mu_j,  M_{[1:j-1}])^2 }{J_{.,opt,j-1}}
  P(d \bmu_k)\dots P(d\bmu_1 )
| 
$$
For a sufficiently large $m$ 
$\prod_{j=2}^{k} \frac{D(  \mu_j,  M_{[1:j-1}])^2 }{J_{m,opt,j-1}}$
differs from 
$\prod_{j=2}^{k} \frac{D(  \mu_j,  M_{[1:j-1}])^2 }{J_{.,opt,j-1}}$ 
by at most $\epsilon_{opt}$. 
Hence 
$$
\le
|
\prod_{j=2}^{k} \frac1{J_{.,opt,j-1}}
  \int_{M\in \textbf{M}_\delta}  J_{cp}(M,P_m)\prod_{j=2}^{k} 
{D(  \mu_j,  M_{[1:j-1}])^2 }
   P_m(d \bmu_k) \dots P_m(d \bmu_1)
$$ $$ %
 -
 (\prod_{j=2}^{k} \frac1{J_{.,opt,j-1}}+/-\epsilon_{opt})
 \int_{M\in \textbf{M}_\delta}  J_{cp}(M,P )\prod_{j=2}^{k}  {D(  \mu_j,  M_{[1:j-1}])^2 }
  P(d \bmu_k)\dots P(d\bmu_1 )
| 
$$
$$
\le
\epsilon_{opt}|
  \int_{M\in \textbf{M}_\delta}  J_{cp}(M,P) \prod_{j=2}^{k} {D(  \mu_j,  M_{[1:j-1}])^2 }
  P(d \bmu_k)\dots P(d\bmu_1 )
|
$$ $$+ \prod_{j=2}^{k} \frac1{J_{.,opt,j-1}} 
|  \int_{M\in \textbf{M}_\delta}  J_{cp}(M,P_m) \prod_{j=2}^{k} {D(  \mu_j,  M_{[1:j-1}])^2 }
  P_m(d \bmu_k) \dots P_m(d \bmu_1)
$$ $$ %
 -
 \int_{M\in \textbf{M}_\delta}  J_{cp}(M,P ) \prod_{j=2}^{k} {D(  \mu_j,  M_{[1:j-1}])^2 }
  P(d \bmu_k)\dots P(d\bmu_1 )
| 
$$
$$
=
\epsilon_{opt}|
  \int_{M\in \textbf{M}_\delta}  J_{cp}(M,P) \prod_{j=2}^{k} {D(  \mu_j,  M_{[1:j-1}])^2 }
  P(d \bmu_k)\dots P(d\bmu_1 )
|
$$ $$+ \prod_{j=2}^{k} \frac1{J_{.,opt,j-1}} 
|  \int_{M\in \textbf{M}_\delta} \int_{\textbf{x}} D(\textbf{x},M)^2   \prod_{j=2}^{k} {D(  \mu_j,  M_{[1:j-1}])^2 }
P_m(d \text{x})    P_m(d \bmu_k) \dots P_m(d \bmu_1)
$$ $$ %
 -
 \int_{M\in \textbf{M}_\delta} \int_{\textbf{x}} D(\textbf{x},M)^2   \prod_{j=2}^{k} {D(  \mu_j,  M_{[1:j-1}])^2 }
P(d \text{x})    P(d \bmu_k)\dots P(d\bmu_1 ) 
| 
$$

The first summand, the product of an $\epsilon_{opt}$ and a finite quantity, can be set as low as needed by choosing a sufficiently low  $\epsilon_{opt}$.

Now following Pollard, we will decrease as much as required the second summand. 
 
So select a finite set $T_\beta$  of points from 
 $\textbf{M}_\delta $ such that 
each element of 
 $\textbf{M}_\delta  $  
lies within a distance of $\beta$ from a point of 
 $T_\beta$. 

Let us  define the function $g_M(x)=  D(\textbf{x},M)^2   \prod_{j=2}^{k} {D(  \mu_j,  M_{[1:j-1}])^2 }$. 

Let us introduce also 
$\overline{D}(x,M)$ and $\underline{D}(x,M)$ as follows:
$M^*$ be the set of elements from $T_\beta$ such that $\|\mu_j-M^*[j]\|$ is less than  $\beta$ . 
then 
$\overline{D}(x,M)=D(x,M^*)+\beta$,
$\underline{D}(x,M)=\max(D(x,M^*)-\beta,0)$.
Now define the function $$\underline{g}_M(x)=  \underline{D}(\textbf{x},M)^2   \prod_{j=2}^{k} {\underline{D}(  \mu_j,  M_{[1:j-1}])^2 }$$
and
 $$\overline{g}_M(x)=  \overline{D}(\textbf{x},M)^2   \prod_{j=2}^{k} {\overline{D}(  \mu_j,  M_{[1:j-1}])^2 }$$

As   $\|x,M^*[j]\|-\beta \le \|x-\mu_j| \le \|x,M^*[j]\|+\beta $,
it is easily seen that 
$\underline{D}(x,M)\le  {D}(x,M)\le \overline{D}(x,M)$
and hence 
$\underline{g}_M(x)\le  {g}_M(x )\le \overline{g}_M(x)$. 

Therefore 
$$|
 \int_{M\in \textbf{M}_\delta} \int_{\textbf{x}} g_M(x) P(dx)   P(d \bmu_k)\dots P(d\bmu_1 )
)
$$ $$- 
 \int_{M\in \textbf{M}_\delta} \int_{\textbf{x}} g_M(x) P_m(dx)   P(d \bmu_k)\dots P(d\bmu_1 )
|$$
is bounded from above by 
$$ 
 \int_{M\in \textbf{M}_\delta} \int_{\textbf{x}} (\overline{g}_M(x)-\underline{g}_M(x) )  P(dx)   P(d \bmu_k)\dots P(d\bmu_1 )
$$ $$+\max(|
 \int_{M\in \textbf{M}_\delta} \int_{\textbf{x}} \overline{g}_M(x) P(dx)   P(d \bmu_k)\dots P(d\bmu_1 )
$$ $$- 
 \int_{M\in \textbf{M}_\delta} \int_{\textbf{x}} \overline{g}_M(x) P_m(dx)   P(d \bmu_k)\dots P(d\bmu_1 )
|
$$ $$,|
 \int_{M\in \textbf{M}_\delta} \int_{\textbf{x}} \underline{g}_M(x) P(dx)   P(d \bmu_k)\dots P(d\bmu_1 )
$$ $$- 
 \int_{M\in \textbf{M}_\delta} \int_{\textbf{x}} \underline{g}_M(x)  P_m(dx)   P(d \bmu_k)\dots P(d\bmu_1 )
| )
$$
A sufficient increase on $m$ will make the second summand as small as we like. 
We will show below that using an argument similar to Pollard, one can show that the first summand can be diminished as much as we like to by appropriate choice of $\beta$. 
Hence the restricted expectation can be shown to converge as claimed. 

For any $R$ 
$$ 
 \int_{M\in \textbf{M}_\delta} \int_{\textbf{x}} (\overline{g}_M(x)-\underline{g}_M(x) )  P(dx)  P(d \bmu_k)\dots P(d\bmu_1 
$$ $$=
 \int_{M\in \textbf{M}_\delta} \int_{\|\textbf{x}||\le R } (\overline{g}_M(x)-\underline{g}_M(x) )  P(dx)  P(d \bmu_k)\dots P(d\bmu_1 
$$ $$+
 \int_{M\in \textbf{M}_\delta} \int_{\|\textbf{x}|| > R } (\overline{g}_M(x)-\underline{g}_M(x) )  P(dx)  P(d \bmu_k)\dots P(d\bmu_1 
$$

The second summand has the property:
$$ \int_{M\in \textbf{M}_\delta} \int_{\|\textbf{x}|| > R } (\overline{g}_M(x)-\underline{g}_M(x) )  P(dx)   P(d \bmu_k)\dots P(d\bmu_1 )
$$ $$\le
\int_{M\in \textbf{M}_\delta} \int_{\|\textbf{x}|| > R } \overline{g}_M(x)   P(dx)   P(d \bmu_k)\dots P(d\bmu_1 )
$$
$$
\le
\int_{M\in \textbf{M}_\delta} \int_{\|\textbf{x}|| > R } \overline(2R_k+2\sigma)^{2k+2} (\|x\|+R_k+2\sigma)^2    P(dx)   P(d \bmu_k)\dots P(d\bmu_1 )
$$
which can be decreased as much as necessary by increasing $R$. 

The first summand can be transformed 
$$
 \int_{M\in \textbf{M}_\delta} \int_{\|\textbf{x}||\le R } (\overline{g}_M(x)-\underline{g}_M(x) )  P(dx)   P(d \bmu_k)\dots P(d\bmu_1 )
$$ $$\le 
 \int_{M\in \textbf{M}_\delta} \int_{\|\textbf{x}||\le R } \sigma   h_M(x)    P(dx)   P(d \bmu_k)\dots P(d\bmu_1 )
$$
where $h_M(x)$ is a function of $x$ and $M$. 
As both $x$ and elements of $M$ are of bounded length, 
$h_M(x)$ is limited from above. 
Hence this summand can be decreased to the desired value by appropriate choice of (small) $\sigma$. 

Therefore the whole term can be decreased to a desired value (above zero of course), what we had to prove.

Hence our claim about consistency of $k$-means++ has been proven.  

\section{Conclusions}
We have shown that the expected value 
of the realistic  $k$-means++ algorithm for finite sample
 converges towards 
the expected value 
of the realistic  $k$-means++ algorithm for the population, when the sample size increases. 
Together with Pollard's results about convergence of $k$-means-ideal 
and the properties proved by Arthur and Vassilvitskii for $k$-means++ on finite samples 
we know now that $k$-means++ provides in expectation also with constant factor approximation of the population optimal $k$-means cost function.

Underway we have also shown that the sample based centre group distribution comes close to that of the population distribution 
so that the partitions of samples yielding constant approximations of $k$-means cost function come close to partitions yielding constant approximations for the entire population.

\bibliographystyle{plain}
\bibliography{kplusplisconvergence_bib}
\end{document}